\def\year{2021}\relax
\documentclass[letterpaper]{article} 
\usepackage{aaai21}  
\usepackage{times}  
\usepackage{helvet} 
\usepackage{courier}  
\usepackage[hyphens]{url}  
\usepackage{graphicx} 
\usepackage{graphicx}  
\usepackage{natbib}  
\usepackage{caption} 
\usepackage[nottoc]{tocbibind}
\usepackage{amsmath}
\usepackage{mathtools}

\usepackage[utf8]{inputenc} 
\usepackage[T1]{fontenc}    
\usepackage{url}            
\usepackage{booktabs}       
\usepackage{amsfonts}       
\usepackage{nicefrac}       
\usepackage{microtype}      
\usepackage{multirow}
\usepackage{amssymb}

\newcommand{\realnumbers}{\ensuremath{\mathbb{R}}}
\newcommand{\transp}{\intercal}

\title{TIME: Text and Image Mutual-Translation Adversarial Networks}

\author{
Bingchen Liu, Kunpeng Song, Yizhe Zhu, Gerard de Melo, Ahmed Elgammal\\
Department of Computer Science, Rutgers University\\
{\tt \{bingchen.liu, kunpeng.song, yizhe.zhu\}@rutgers.edu\\
gdm@demelo.org, elgammal@cs.rutgers.edu}
}

\begin{document}
\maketitle
\begin{abstract}
  Focusing on text-to-image (T2I) generation, we propose Text and Image Mutual-Translation Adversarial Networks (TIME), a lightweight but effective model that jointly learns a T2I generator $G$ and an image captioning discriminator $D$ under the Generative Adversarial Network framework. While previous methods tackle the T2I problem as a uni-directional task and use pre-trained language models to enforce the image--text consistency, TIME requires neither extra modules nor pre-training. We show that the performance of $G$ can be boosted substantially by training it jointly with $D$ as a language model. Specifically, we adopt Transformers to model the cross-modal connections between the image features and word embeddings, and design an annealing conditional hinge loss that dynamically balances the adversarial learning. In our experiments, TIME achieves state-of-the-art (SOTA) performance on the CUB and MS-COCO dataset (Inception Score of 4.91 and Fréchet Inception Distance of 14.3 on CUB), and shows promising performance on MS-COCO on image captioning and downstream vision-language tasks. 
\end{abstract}

\section{Introduction}
\label{sec:intro}

There are two main aspects to consider when approaching the text-to-image (T2I) task: the image generation quality and the image--text semantic consistency. The T2I task is commonly modeled by a conditional Generative Adversarial Network (cGAN)~\cite{mirza2014conditional,goodfellow2014generative}, where a Generator ($G$) is trained to generate images given the texts describing the contents, and a Discriminator ($D$) learns to determine the authenticity of the images, conditioned on the semantics defined by the given texts.

To address the first aspect, \citeauthor{zhang2017stackgan} (\citeyear{zhang2017stackgan}) introduced StackGAN by letting $G$ generate images at multiple resolutions, and adopted multiple $D$s to jointly refine $G$ from coarse to fine levels. StackGAN invokes a pre-trained Recurrent-Neural-Network (RNN) \cite{hochreiter1997long,mikolov2010recurrent} to provide text conditioning for the image generation. To approach the second aspect, \citeauthor{xu2018attngan} (\citeyear{xu2018attngan}) take StackGAN as the base model and propose AttnGAN, which incorporates word embeddings into the generation and consistency-checking processes. A pre-trained Deep-Attentional-Multimodal-Similarity-Model (DAMSM) is introduced, which better aligns the image features and word embeddings via attention mechanism.

\begin{figure*}
\centering
  \includegraphics[width=0.8\linewidth,height=0.30\linewidth]{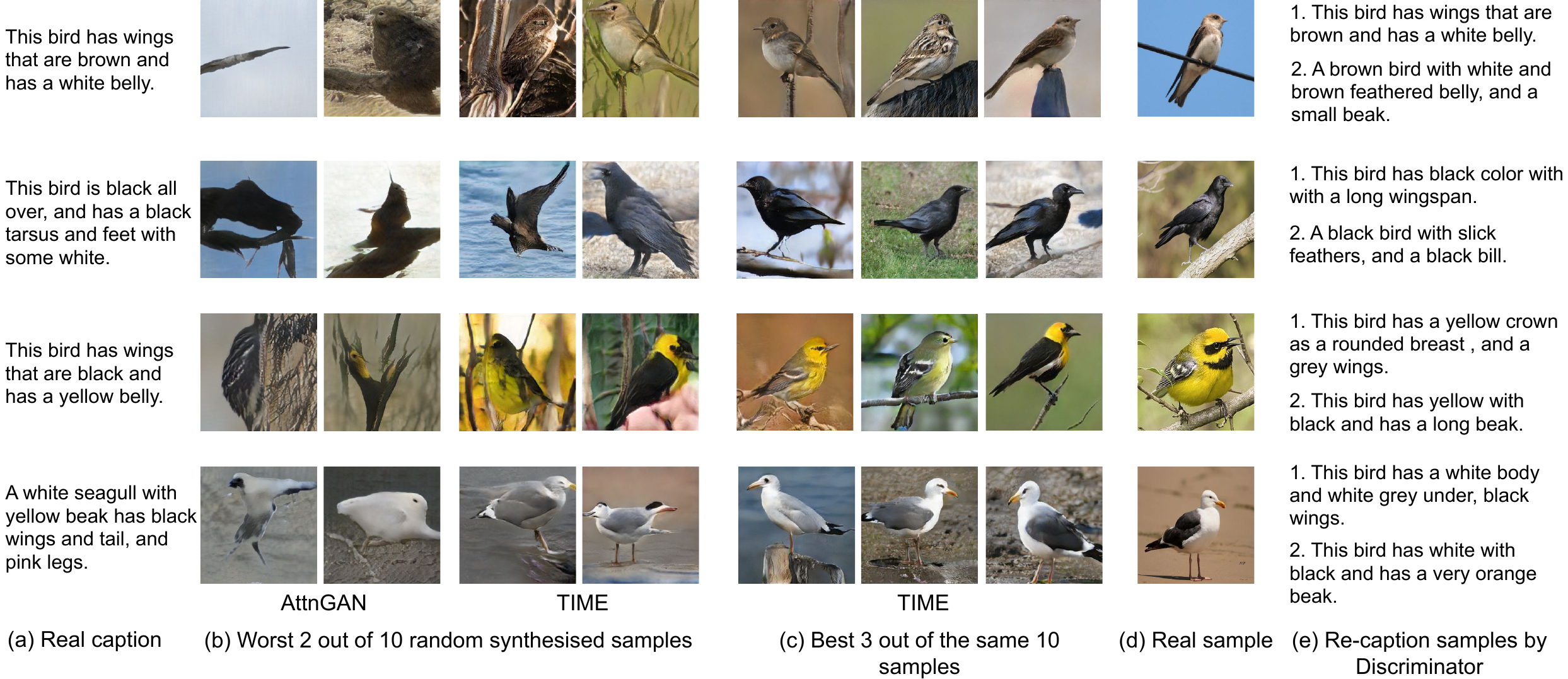}
  \caption{Text-to-image results of TIME on the CUB dataset, where $D$ works as a stand-alone image-captioning model.}
  \label{fig:quality_1}
\end{figure*}

While the T2I performance continues to advance \cite{qiao2019mirrorgan,zhu2019dm,8930532,NIPS2019_8480,yin2019semantics,hinz2019semantic}, the follow-up methods all share two common traits. First, they all adopt the same stacked structure of $G$ that requires multiple $D$s. Second, they all rely on the pre-trained DAMSM from AttnGAN to maintain the image--text consistency. However, these methods fail to take advantage of recent advances in both the GAN and NLP literature \cite{karras2017progressive,karras2019style,vaswani2017attention,devlin2018bert,radford2019language}. The rapidly progressing research in these two fields provides the opportunity to explore a substantial departure from previous works on text-to-image modeling. In particular, as StackGAN and follow-up works all depend on: 1. a pre-trained text encoder for word and sentence embeddings; 2. an additional image encoder to ascertain image--text consistency; two important questions arise. First, can we skip the pre-training step and elegantly train the text encoder as part of $D$? Second, can we abandon the extra CNN (in the DAMSM module which extracts image features) and use $D$ as the image encoder? If the answers are affirmative, two further questions can be explored. When $D$ and the text encoder are jointly trained to match the visual and text features, can we obtain an image captioning model from them? Furthermore, since $D$ is trained to extract text-relevant image features, will it benefit $G$ in generating more semantically consistent images? 

With these questions in mind, we present the Text and Image Mutual-translation adversarial nEtwork (TIME). To the best of our knowledge, this is the first work that jointly handles both text-to-image and image captioning in a single model using the GAN framework. Our contributions can be summarized as follows:
\begin{enumerate}
    \item We propose an efficient model: Text and Image Mutual-Translation Adversarial Networks (TIME), for T2I tasks trained in an end-to-end fashion, without any need for pre-trained models or complex training strategies.
    \item We introduce two techniques: 2-D positional encoding for a better attention operation and annealing hinge loss to dynamically balance the learning paces of $G$ and $D$. 
    \item We show that the sentence-level text features are no longer needed in T2I task, which leads to a more controllable T2I generation that is hard to achieve in previous models.
    \item Extensive experiments show that our proposed TIME achieves  superior results on text-to-image tasks and promising results on image captioning. Fig.~\ref{fig:quality_1}-(c) showcases the superior synthetic image quality from TIME, while Fig.~\ref{fig:quality_1}-(e) demonstrates TIME's image captioning capability.
\end{enumerate}

\section{Related Work and Background}
\label{sec:related}
Recent years have witnessed substantial progress in the text-to-image task \cite{mansimov2015generating,nguyen2017plug,reed2017parallel,reed2016generative,zhang2017stackgan,xu2018attngan,han2019art} owing largely to the success of deep generative models~\cite{goodfellow2014generative,kingma2013auto,van2016conditional}. \citeauthor{reed2016generative} first demonstrated the superior ability of conditional GANs to synthesize plausible images from text descriptions. StackGAN and AttnGAN then took the generation quality to the next level, which subsequent works built on \cite{qiao2019mirrorgan,zhu2019dm,8930532,NIPS2019_8480,yin2019semantics,hinz2019semantic,li2019object} . Specifically, MirrorGAN \cite{qiao2019mirrorgan} incorporates a pre-trained text re-description RNN to better align the images with the given texts, DMGAN \cite{zhu2019dm} integrates a dynamic memory module on $G$, ControlGAN~\cite{NIPS2019_8480} employs a channel-wise attention in $G$, and SDGAN~\cite{yin2019semantics} includes a contrastive loss to strengthen the image--text correlation. In the following, we describe the key components of StackGAN and AttnGAN.

\textbf{StackGAN as the Image Generation Backbone.} StackGAN adopts a coarse-to-fine structure that has shown substantial success on the T2I task. In practice, the generator $G$ takes three steps to produce a $256\times256$ image, where three discriminators ($D$) are required to train $G$. However, a notable reason for seeking an alternative architecture is that the multi-$D$ design is memory-demanding and has a high computational burden during training. If the image resolution increases, the respective higher-resolution $D$s can raise the cost particularly dramatically. 

\textbf{Dependence on Pre-trained modules.} While the overall framework for T2I models resembles a conditional GAN (cGAN), multiple modules have to be pre-trained in previous works. In particular, AttnGAN requires a DAMSM, which includes an Inception-v3 model \cite{szegedy2016rethinking} that is first pre-trained on ImageNet \cite{deng2009imagenet}, and then used to pre-train an RNN text encoder. MirrorGAN further proposes the STREAM model, which is also an additional CNN+RNN structure pre-trained for image captioning.

Such pre-training has several drawbacks, including, first, the additional pre-trained CNN for image feature extraction introduces a significant amount of weights, which can be avoided as we shall later show. Second, using pre-trained modules leads to extra hyper-parameters that require dataset-specific tuning. For instance, in AttnGAN, the weight for the DAMSM loss can range from 0.2 to 100 across  different datasets. Last but not least, empirical studies \cite{qiao2019mirrorgan,zhang2017stackgan} show that the pre-trained NLP components do not converge if jointly trained with the cGAN.

\textbf{The Image-Text Attention Mechanism.} The attention mechanism employed in AttnGAN can be interpreted as a simplified version of the Transformer \cite{vaswani2017attention}, where the three-dimensional image features (height$\times$width$\times$channel) in the CNN are flattened into a two-dimensional sequence (seq-length$\times$channel where seq-length$=$height$\times$width). This process is demonstrated in Fig.~\ref{fig:attn_compare}-(a), where an image-context feature $f_\mathrm{it}$ is derived via an attention operation on the reshaped image feature and the sequence of word embeddings. The resulting image-context features are then concatenated to the image features to generate the images. We will show that a full-fledged version of the Transformer can further improve the performance without a substantial additional computational burden.

\begin{figure*}[h]
\centering
  \includegraphics[width=0.82\linewidth,height=0.34\linewidth]{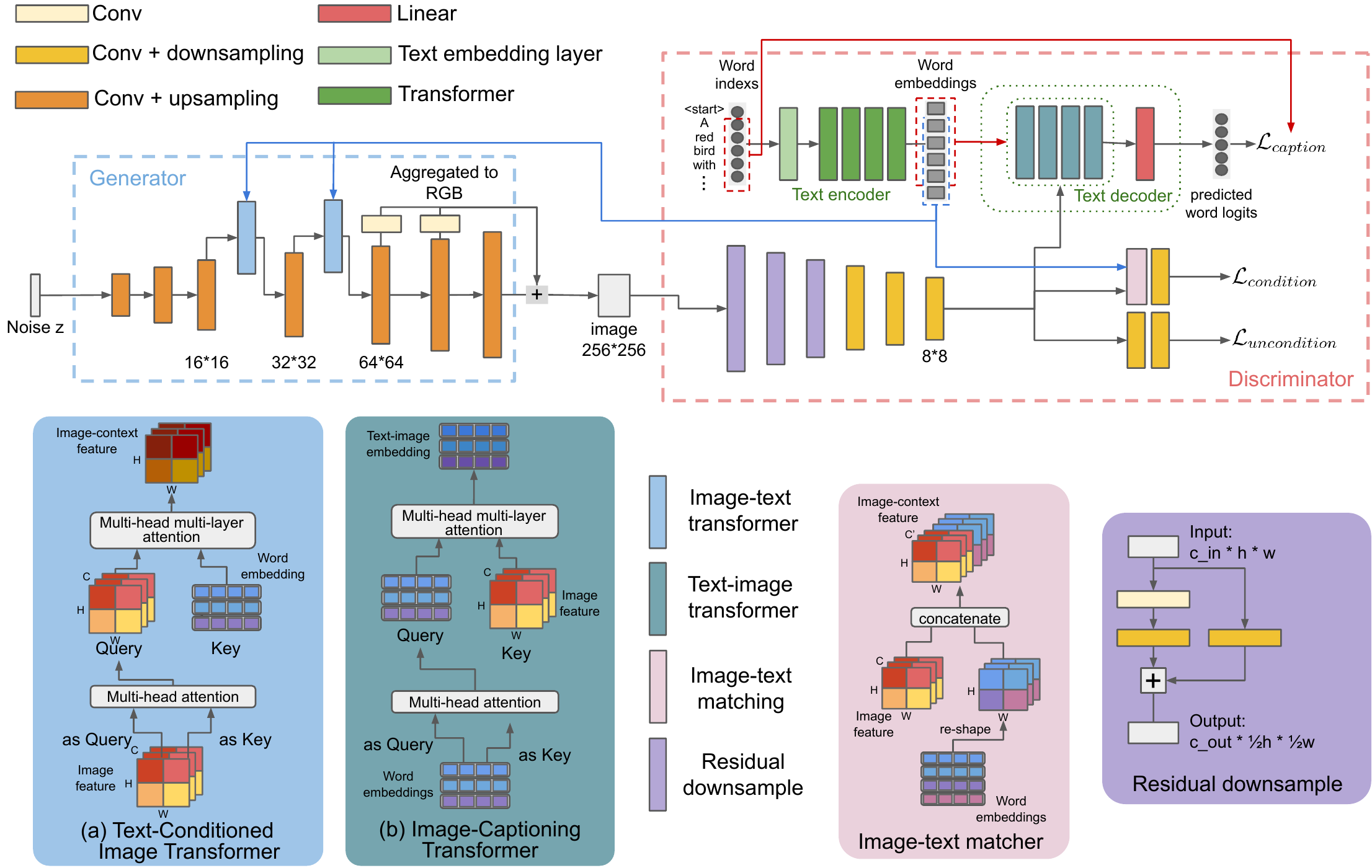}
  \caption{Model overview of TIME. The upper panel shows  a high-level summary of our architecture while the lower panel demonstrates the details of the individual modules.}
  \label{fig:model_overview}
\end{figure*}

\section{The Motivation of Mutual Translation}
One may ask that since training the text-to-image model already achieves fruitful results with a pre-trained NLP model, is it necessary to explore the joint-training method? We can answer this question from several aspects. 

First, a pre-trained NLP model is not always available given some image datasets. In cases where the given texts do not have a pre-trained NLP model, one can save the separated pre-training time and get a model that translates towards both directions with TIME. In case a pre-trained NLP model is available, it is still not guaranteed that the fixed word embeddings are the best for training the image generator. Tuning the hyper-parameters (such as weights of loss objectives from the pre-trained NLP model) for the pre-training methods can be very costly and may not be optimal.

Second, under the GAN framework, balancing the joint training between the Discriminator $D$ and Generator $G$ is vital. $G$ is unlikely to converge if trained with a fixed $D$. In the text-to-image task, the pre-trained NLP model serves as a part of $D$ that provides authenticity signals to $G$. Using a pre-trained NLP model is equivalent to fixing a part of $D$, which undermines the performance of the whole training schema as a GAN. Instead, the joint training in TIME does not have such restrictions. The NLP parts in TIME learn together with $G$ and dynamically adjust the word embeddings to serve the training best, leading to better image synthesis quality. 

Finally, mutual translation itself can be a crucial pre-training method, which is also studied in \cite{huang2018turbo,li2020oscar}. As we show in the paper, the NLP models learned in TIME have a promising performance on downstream vision-language tasks. Therefore, instead of pre-training only on texts, mutual translation between image and text itself has the potential to be a powerful pre-training method.  

\section{Methodology}

\label{sec:TIME}
In this section, we present our proposed approach. The upper panel in Fig.~\ref{fig:model_overview} shows the overall structure of TIME, consisting of a Text-to-Image Generator $G$ and an Image-Captioning Discriminator $D$. We treat a text encoder $Enc$ and a text decoder $Dec$ as parts of $D$. $G$'s Text-Conditioned Image Transformer accepts a series of word embeddings from $Enc$ and produces an image-context representation for $G$ to generate a corresponding image. $D$ is trained on three kinds of input pairs, consisting of captions $T^\mathrm{real}$ alongside: (a) matched real images $I_\mathrm{match}$; (b) randomly mismatched real images $I_\mathrm{mis}$; and (c) generated images $I_\mathrm{fake}$ from $G$.

\subsection{Model Structures}
\subsubsection{Aggregated Generator} To trim the model size of the StackGAN structure, we present the design of an aggregated $G$ as shown in the upper panel of Fig.~\ref{fig:model_overview}. $G$ still yields RGB outputs at multiple resolutions, but these RGB outputs are re-scaled and added together as a single aggregated image output. Therefore, only one $D$ is needed to train $G$. 

\label{sec:backbone-attn}
\subsubsection{Text-Conditioned Image Transformer} 
\begin{figure*}
\centering
  \includegraphics[width=0.9\linewidth,height=0.28\linewidth]{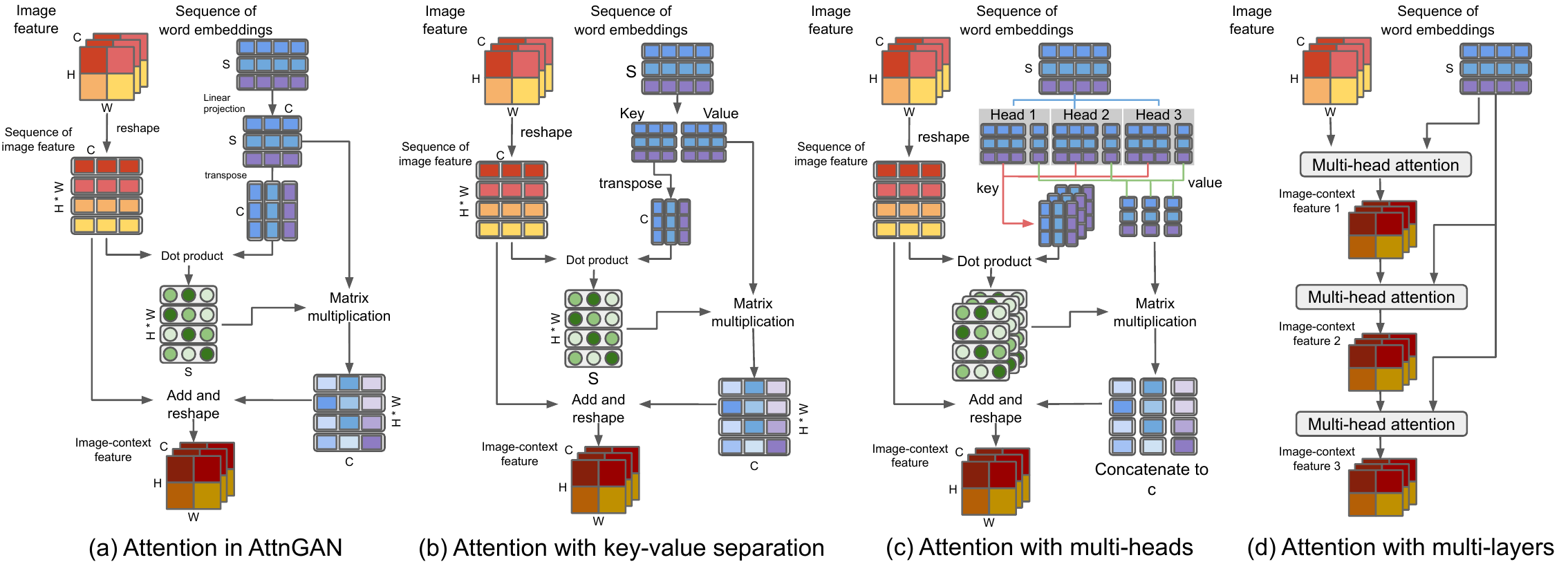}
  \caption{Differences between the attention of AttnGAN and TCIT.}
  \label{fig:attn_compare}
\end{figure*}

While prior works \cite{zhang2018self,xu2018attngan} show the benefit of an attention mechanism for the image generative task, none of them dive deeper towards the more comprehensive ``multi-head and multi-layer" Transformer design \cite{vaswani2017attention}. To explore a better baseline for the T2I task, we redesign the attention in AttnGAN with the Text-Conditioned Image Transformer (TCIT) as illustrated in Fig.~\ref{fig:model_overview}-(a). In Fig.~\ref{fig:attn_compare}, we show three main differences between TCIT and the form of attention that is widely used in previous T2I models such as AttnGAN. All the attention modules take two inputs: the image feature $f_i$, and the sequence of word embeddings $f_t$, and gives one output: the revised image feature $f_{it}$ according to the word embeddings $f_t$. 

First, Fig.~\ref{fig:attn_compare}-(a) shows the attention module from AttnGAN, where the projected key ($K$) from $f_\mathrm{t}$ is used for both matching with query ($Q$) from $f_i$ and calculating $f_\mathrm{it}$. Instead, TCIT has two separate linear layers to project $f_\mathrm{t}$ as illustrated in Fig.~\ref{fig:attn_compare}-(b). The intuition is, as $K$ focuses on matching with $f_\mathrm{i}$, the other projection value $V$ can better be optimized towards refining $f_\mathrm{i}$ for a better $f_\mathrm{it}$. Second, TCIT adopts a multi-head structure as shown in Fig.~\ref{fig:attn_compare}-(c). Unlike in AttnGAN where only one attention map is applied, the Transformer replicates the attention module, thus adding more flexibility for each image region to account for multiple words. Third, TCIT stacks the attention layers in a residual structure as in certain NLP models \cite{devlin2018bert,radford2019language} as illustrated in Fig.~\ref{fig:attn_compare}-(d), for a better performance by provisioning multiple attention layers and recurrently revising the learned features. In contrast, previous GAN models (AttnGAN, SAGAN) adopt attention only in a one-layer fashion.

\subsubsection{Image-Captioning Discriminator}
\label{sec:baseline-TIME}
We treat the text encoder $Enc$ and text decoder $Dec$ as a part of our $D$. Specifically, $Enc$ is a Transformer that maps the word indices into the embeddings while adding contextual information to them. To train $Dec$ to actively generate text descriptions of an image, an attention mask is applied on the input of $Enc$, such that each word can only attend to the words preceding it in a sentence. $Dec$ is a Transformer decoder that performs image captioning by predicting the next word's probability from the masked word embeddings and the image features.

\subsubsection{Image-Captioning Transformer} Symmetric to TCIT, the inverse operation, in which $f_\mathrm{t}$ is revised by $f_\mathrm{i}$, is leveraged for image captioning in $Dec$, as shown in Fig.~\ref{fig:model_overview}-(b). Such a design has been widely used in recent captioning works. In TIME, we show that a simple 4-layer 4-head Transformer is sufficient to obtain high-quality captions and facilitate the consistency checking in the T2I task. 

\begin{figure}[h]
\centering
  \includegraphics[width=0.9\linewidth]{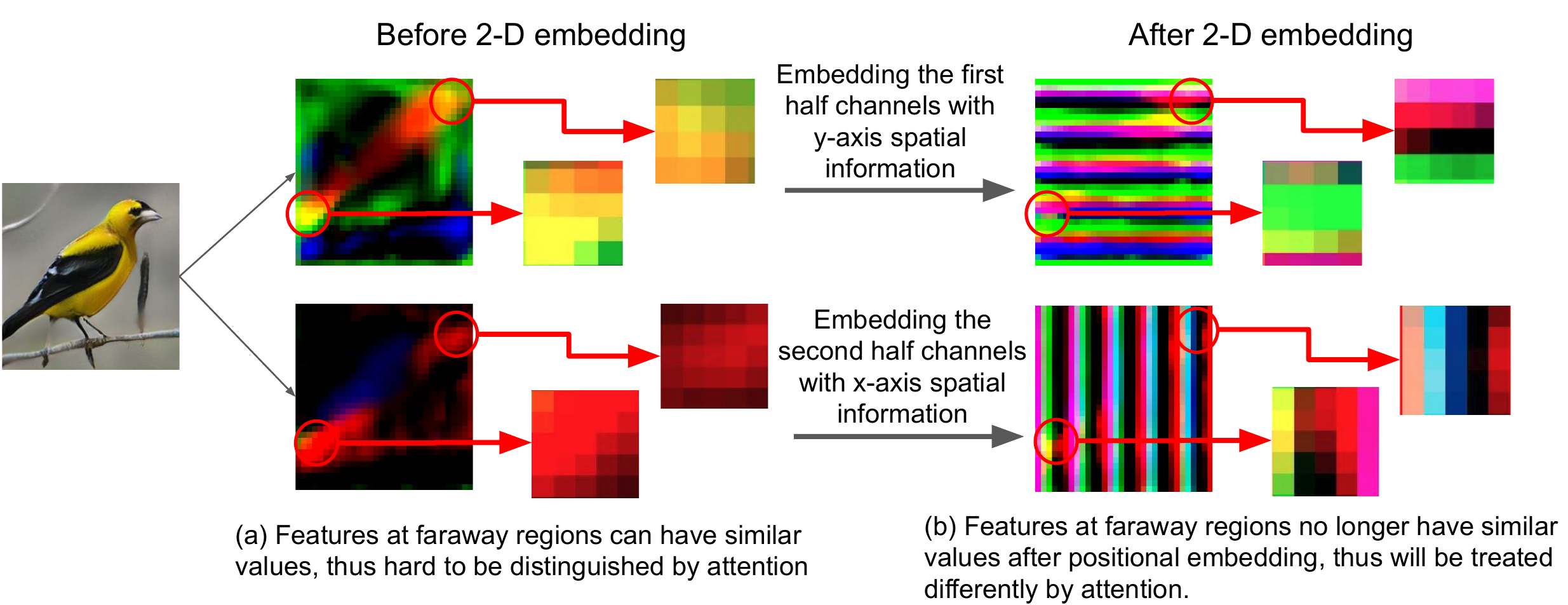}
  \caption{Visualization of 2-D positional embedding.}
  \label{fig:2-d-positional-embedding}
\end{figure}

\subsection{2-D Positional Encoding for Image Features}

When we reshape the image features $f_i$ for the attention operation, there is no way for the Transformer to discern spatial information from the flattened features. To take advantage of coordinate signals, we propose \emph{2-D positional encoding} as a counter-part to the 1-D \emph{positional encoding} in the Transformer \cite{vaswani2017attention}. The encoding at each position has the same dimensionality as the channel size $c$ of $f_i$, and is directly added to the reshaped image feature $f_\mathrm{i}^\transp \in \realnumbers^{d \times c}$. The first half of dimensions encode the y-axis positions and the second half encode the x-axis, with sinusoidal functions of different frequencies. Such 2-D encoding ensures that closer visual features have a more similar representation compared to features that are spatially more remote from each other. An example $32\times32$ feature-map from a trained TIME is visualized in Fig.~\ref{fig:2-d-positional-embedding}, where we visualize three feature channels as an RGB image. In practice, we apply 2-D positional encoding on the image features for both TCIT and $Dec$ in $D$. Please refer to the appendix for more details.

\subsection{Objectives}

\subsubsection{Discriminator Objectives}
Formally, we denote the three kinds of outputs from $D$ as: $D_\mathrm{f}()$, the image feature at $8 \times 8$ resolution; $D_\mathrm{u}()$, the unconditional image real/fake score; and $D_\mathrm{c}()$, the conditional image real/fake score. Therefore, the predicted next word distribution from $Dec$ is: $P_\mathrm{k} = Dec( Enc(T^\mathrm{real}_\mathrm{1:k-1} ), D_\mathrm{f}(I_\mathrm{match}))$. Finally, the objectives for $D, Enc$, and $Dec$ to jointly minimize are:
\begin{align}
    \mathcal{L}_\mathrm{caption} =& - \sum_\mathrm{k=1}^{l} \log(P_\mathrm{k}(T^\mathrm{real}_\mathrm{k}, D_\mathrm{f}(I_\mathrm{match}))) ;\\
    \nonumber
    \mathcal{L}_\mathrm{uncond} =& - \mathbb{E}[\log(D_\mathrm{u}(I_\mathrm{match}))] \\ 
    &- \mathbb{E}[\log(1 - D_\mathrm{u}(I_\mathrm{fake}))] ;
\end{align}
along with $\mathcal{L}_\mathrm{cond}$, which we shall discuss next.

\begin{figure}
    \centering
   \includegraphics[width=0.9\linewidth]{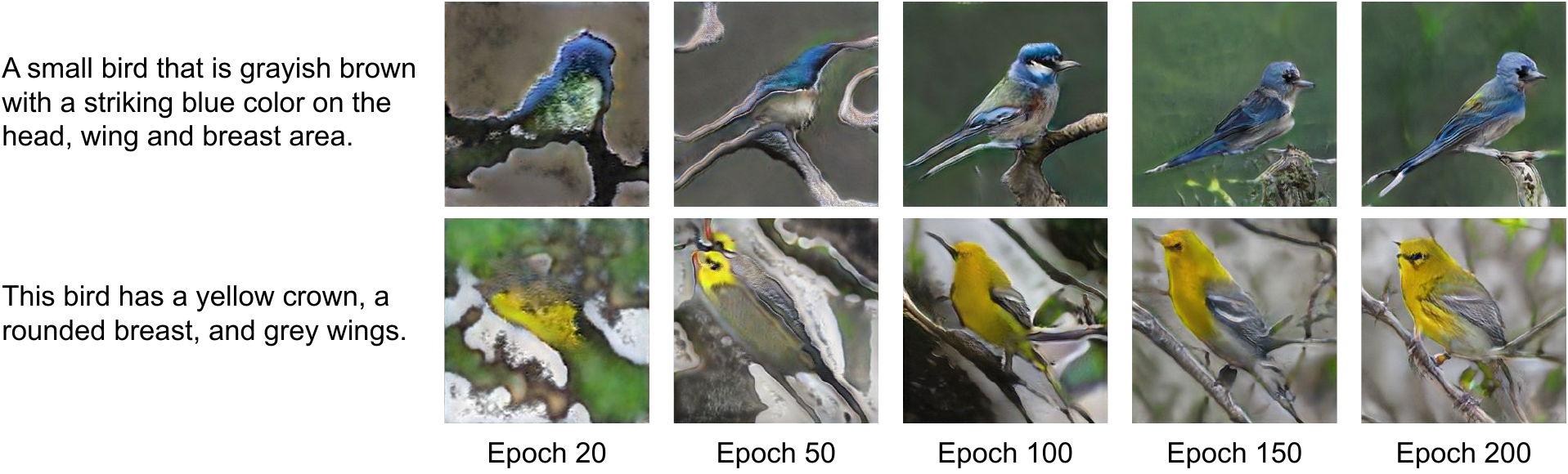}
  \caption{Samples generated during the training of TIME, note the visual features emerge in very early iterations.}
  \label{fig:hinge}
\end{figure}

\subsubsection{Annealing Image--Text Matching Loss}
During training, we find that $G$ can learn a good semantic visual translation at very early iterations. As shown in Fig.~\ref{fig:hinge}, while the convention is to train the model for 600 epochs on the CUB dataset, we observe that the semantic features of $T^\mathrm{real}$ begin to emerge on $I_\mathrm{fake}$ as early as after 20 epochs. Thus, we argue that it is not ideal to penalize $I_\mathrm{fake}$ by the conditional loss on $D$ in a static manner. Since $I_\mathrm{fake}$ is already very consistent to the given $T^\mathrm{real}$, if we let $D$ consider an already well-matched input as inconsistent, this may confuse $D$ and in turn hurt the consistency-checking performance. Therefore, we employ a hinge loss \cite{lim2017geometric,tran2017deep} and dynamically anneal the penalty on $I_\mathrm{fake}$ according to how confidently $D$ predicts the matched real pairs:
\begin{align}
&s_\mathrm{pivot} = \mathrm{detach}(\mathbb{E}[D_\mathrm{c}(I_\mathrm{match}, Enc(T^\mathrm{real}))]) ;\\
&\mathcal{L}_\mathrm{cond} = \mathbb{E}[\min(0, 1 - D_\mathrm{c}(I_\mathrm{match}, Enc(T^\mathrm{real})))] \nonumber\\
                                & + \mathbb{E}[\min(0, 1 + D_\mathrm{c}(I_\mathrm{mismatch}, Enc(T^\mathrm{real})))] \nonumber\\
                                & + \mathbb{E}[\min(0, -s_\mathrm{pivot} \times p + D_\mathrm{c}(I_\mathrm{fake}, Enc(T^\mathrm{real})))].
\end{align}
Here, $\mathrm{detach}(.)$ denotes that the gradient is not computed for the enclosed function, and $p=i_\mathrm{epoch}/n_\mathrm{epochs}$ (current epoch divided by total number) is the annealing factor. The hinge loss ensures that $D$ yields a lower score on $I_\mathrm{fake}$ compared to $I_\mathrm{match}$, while the annealing term $p$ ensures that $D$ penalizes $I_\mathrm{fake}$ sufficiently in early epochs.

\subsubsection{Generator Objectives} On the other side, $G$ considers random noise $z$ and word embeddings from $Enc$ as inputs, and is trained to generate images that can fool $D$ into giving high scores on authenticity and semantic consistency with the text. Moreover, $G$ is also encouraged to make $D$ reconstruct the same sentences as provided as input. Thus, the objectives for $G$ to minimize are:
\begin{align}
    &\mathcal{L}_\mathrm{caption-g} = - \sum_\mathrm{k=1}^{l} \log(P_\mathrm{k}(T^\mathrm{real}_\mathrm{k}, D_\mathrm{f}( G(z, Enc(T^\mathrm{real})) ))) ;\\
    &\mathcal{L}_\mathrm{uncond-g} = - \mathbb{E}[\log(D_\mathrm{u}( G(z, Enc(T^\mathrm{real})) ))] ;\\
    &\mathcal{L}_\mathrm{cond-g} = - \mathbb{E}[D_\mathrm{c}(G(z, Enc(T^\mathrm{real})), Enc(T^\mathrm{real}))].
\end{align}

\section{Experiments}
In this section, we evaluate the proposed model from both the text-to-image and image-captioning directions, and analyze each module's effectiveness individually. Moreover, we highlight the desirable property of TIME being a more controllable generator compared to other T2I models.  

Experiments are conducted on two datasets: CUB \cite{WelinderEtal2010} and MS-COCO \cite{lin2014microsoft}. We follow the same convention as in previous T2I works to split the training/testing set. We benchmark the image quality by the Inception Score (IS) \cite{salimans2016improved} and Fréchet Inception Distance (FID) \cite{heusel2017gans}, and measure the image--text consistency by R-precision \cite{xu2018attngan} and SOA-C \cite{hinz2019semantic}. 

\subsubsection{Backbone Model Structure}

Table~\ref{table:stack} demonstrates the performance comparison between the StackGAN structure and our proposed ``aggregating" structure. AttnGAN as the T2I backbone has been revised by recent advances in the GAN literature \cite{zhu2019dm,qiao2019mirrorgan}. Seeking a better baseline, we also incorporated recent advances. In particular, columns with ``+new" imply that we train the model with equalized learning rate \cite{karras2017progressive} and R-1 regularization \cite{mescheder2018training}. ``Aggr" means we replace the ``stacked" $G$ and multiple $D$s with the proposed aggregated $G$ and a single $D$. To show the computing cost, we list the relative training times of all models with respect to StackGAN. All models are trained with the optimal hyper-parameter settings and the same loss functions from StackGAN and AttnGAN respectively.

\begin{table}[h]
  \begin{center}
    \resizebox{0.9\linewidth}{!}{
      \begin{tabular}{l|c c c}
            \toprule
            & Inception Score $\uparrow$ & R-precision $\uparrow$ & Training time $\downarrow$ \\
            \cmidrule{1-4}
            StackGAN w/o stack &  $3.42 \pm 0.05$  &  $9.25 \pm 3.12$  &  0.57 \\
            StackGAN &  $3.82 \pm 0.06$  &  $10.37 \pm 5.88$  &  1.0 \\
            Aggr GAN &  $3.78 \pm 0.03$  &  $10.21 \pm 5.42$  & 0.78     \\
            Aggr GAN +new &  $4.12 \pm 0.03$  &  $12.26 \pm 4.76$  & 0.85     \\
            \cmidrule{1-4}
            AttnGAN w/o stack &   $4.28 \pm 0.02$ &  $64.82 \pm 4.43$  &  0.71   \\
            AttnGAN &  $4.36 \pm 0.03$   &  $67.82 \pm 4.43$  &  1.14     \\
            Aggr AttnGAN &  $4.34 \pm 0.02$   &  $67.63 \pm 5.43$  &  0.86     \\
            Aggr AttnGAN +new   &  $\textbf{4.52} \pm 0.02$     &  $\textbf{70.32} \pm 4.19$   &  1.0    \\
            \bottomrule
      \end{tabular}
    }
  \end{center}
\caption{Comparison between \textbf{stacked} and \textbf{aggregated} model structures on the CUB dataset.}
\label{table:stack}
\end{table}

The aggregated structure with new GAN advances achieve the best performance/compute-cost ratio in both the image quality and the image--text consistency. Moreover, we find that the abandoned lower-resolution $D$s in StackGAN have limited effect on image--text consistency, which instead appears more related to the generated image quality, as a higher IS always yields a better R-precision.
\subsubsection{Attention Mechanisms}

We conduct experiments to explore the best attention settings for the T2I task from the mechanisms discussed in Section~\ref{sec:backbone-attn}. 
 
\begin{table}[h]
  \begin{center}
  \resizebox{0.7\linewidth}{!}{
    \begin{tabular}{l | c c }
           \toprule
            & Inception Score $\uparrow$ & R-precision $\uparrow$ \\
           \cmidrule{1-3}
           AttnGAN & $4.36 \pm 0.03$ & $67.82 \pm 4.43$ \\
           Tf-h1-l1 & $4.38 \pm 0.06$  & $66.96 \pm 5.21$  \\
           Tf-h4-l1 & $4.42 \pm 0.06$  &  $68.58 \pm 4.39$ \\
           Tf-h4-l2 & $\textbf{4.48} \pm 0.03$  &  $\textbf{69.72} \pm 4.23$ \\
           Tf-h4-l4 & $4.33 \pm 0.02$   & $67.42 \pm 4.31$  \\
           Tf-h8-l4 & $4.28 \pm 0.03$  &  $62.32 \pm 4.25$  \\
           \bottomrule
           \end{tabular}}
   \end{center}
   \caption{Comparison of different attention settings on CUB.}
   \label{table:attn}
\end{table}

Table~\ref{table:attn} lists the settings we tested, where all the models are configured the same based on AttnGAN, except for the attention mechanisms used in $G$. In particular, column 1 shows the baseline performance that employs the basic attention operation, described in Fig.~\ref{fig:attn_compare}-(a), from AttnGAN. The following columns show the results of using the Transformer illustrated in Fig.~\ref{fig:attn_compare}-(d) with different numbers of heads and layers (e.g., Tf-h4-l2 means a Transformer with 4 heads and 2 layers). The results suggest that a Transformer with a more comprehensive attention yields a better performance than the baseline. However, when increasing the number of layers and heads beyond a threshold, a clear performance degradation emerges on the CUB dataset. More discussion and the results on MS COCO can be found in the appendix.

\begin{figure*}[h]
    \centering
   \includegraphics[width=0.86\linewidth,height=0.25\linewidth]{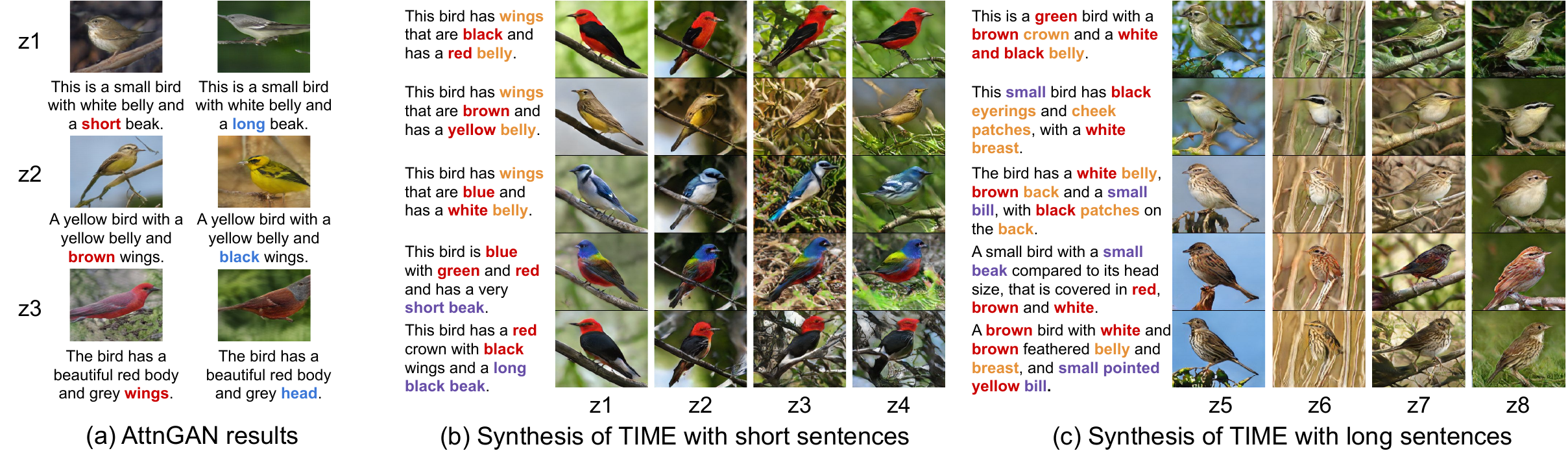}
  \caption{Images from TIME with fixed $z$ and varied sentences}
  \label{fig:quality_2}
\end{figure*}

\subsubsection{Controllable G without Sentence-Embedding} Most previous T2I models rely on a sentence-level embedding $f_\mathrm{s}$ as a vital conditioning factor for $G$ \cite{zhang2017stackgan,xu2018attngan,qiao2019mirrorgan,zhu2019dm,NIPS2019_8480}. Specifically, $f_\mathrm{s}$ is concatenated with noise $z$ as the input for $G$, and is leveraged to compute the conditional authenticity of the images in $D$. Sentence embeddings are preferred over word embeddings, as the latter lack contextual meaning and semantic concepts are often expressed in multiple words.

However, since $f_\mathrm{s}$ is a part of the input alongside $z$, any slight changes in $f_\mathrm{s}$ can lead to major visual changes in the resulting images, even when $z$ is fixed. This is undesirable when we like the shape of a generated image but want to slightly revise it by altering the text description. Examples are given in Fig.~\ref{fig:quality_2}-(a), where changing just a single word leads to unpredictably large changes in the image. In contrast, since we adopt the Transformer as the text encoder, where the word embeddings already come with contextual information, $f_\mathrm{s}$ is no longer needed in TIME. Via our Transformer text encoder, the same word in different sentences or at different positions will have different embeddings. As a result, the word embeddings are sufficient to provide semantic information, and we can abandon the sentence embedding.

In Fig.~\ref{fig:quality_2}-(b) and (c), TIME shows a more controllable generation when changing the captions while fixing $z$. TIME provides a new perspective that naturally enables fine-grained manipulation of synthetic images via their text descriptions.

\subsubsection{Ablation Study}
Based on AttnGAN objectives, we combine the model settings from Table~\ref{table:stack} row.7 and Table~\ref{fig:attn_compare} row.5 as the baseline, and perform ablation study in Table~\ref{table:ablation}. First, we remove the image captioning text decoder $Dec$ to show its positive impact. Then, we add $Dec$ back and show that dropping the sentence-level embedding does not hurt the performance. Adding 2-D positional encoding brings improvements in both image--text consistency and the overall image quality. Lastly, the proposed hinge loss $L_{hinge}$ (eq. 4) releases $D$ from a potentially conflicting signal, resulting in the most substantial boost in image quality. 

\begin{table}[h]
  \begin{center}
    \resizebox{0.9\columnwidth}{!}{
      \begin{tabular}{l| c c }
        \toprule
             &  Inception Score $\uparrow$  &  R-precision $\uparrow$ \\
        \midrule
          Baseline   &  $4.64 \pm 0.03$  &  $70.72 \pm 1.43$   \\
          B - img captioning   & $4.58 \pm 0.02$   &  $69.72 \pm 1.43$   \\
          B - Sentence emb   &  $4.64 \pm 0.06$  &   $68.96 \pm 2.21$  \\
          B + 2D-Pos Encode   &  $4.72 \pm 0.06$  &  $71.58 \pm 2.39$   \\
          B + Hinged loss   &  $4.91 \pm 0.03$   &  $71.57 \pm 1.23$   \\
          \bottomrule
       \end{tabular}
    }
   \end{center}
   \caption{Ablation Study of TIME on CUB dataset}
   \label{table:ablation}
\end{table}

\begin{figure}[h]
    \centering
    \includegraphics[width=0.9\linewidth]{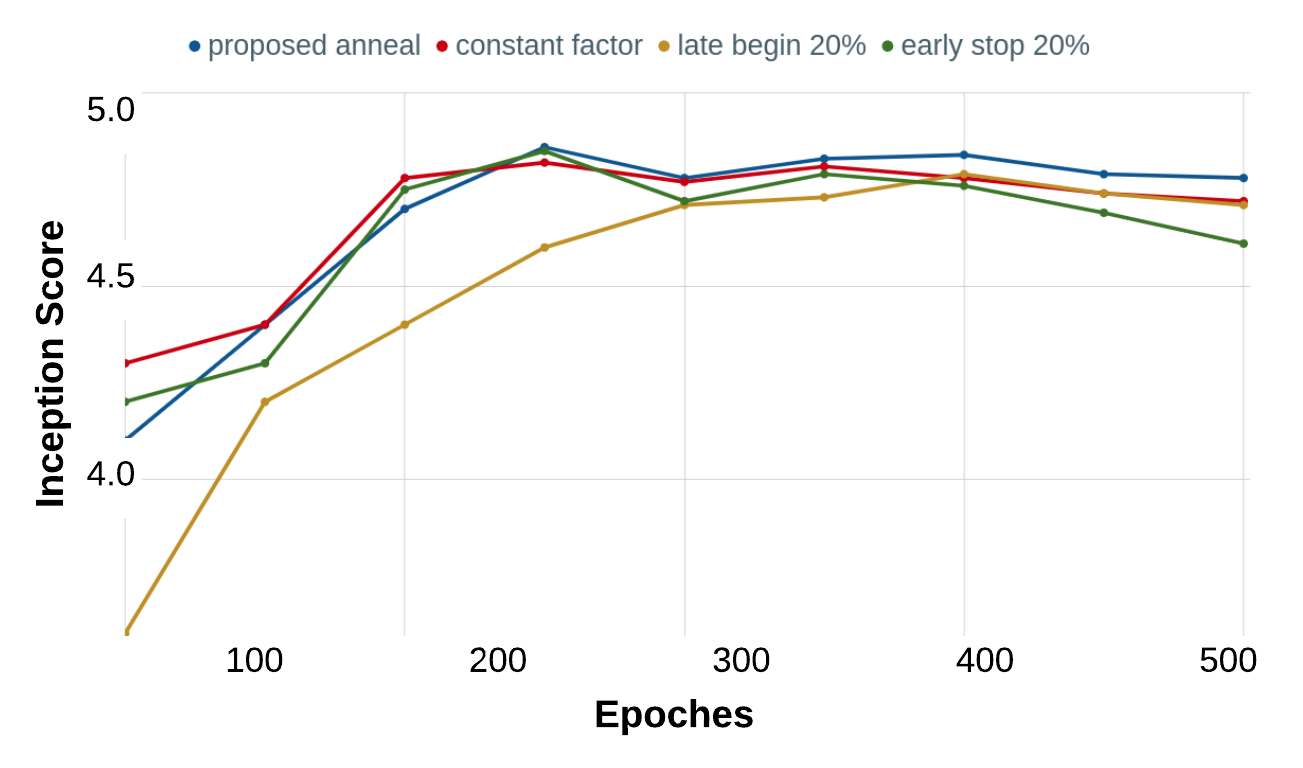}
    \caption{Performance comparison on different annealing schedules of the hinged image-text consistency loss.}
    \label{fig:hingeloss_schedule}
\end{figure}

To emphasize the contribution of the proposed image-text hinge loss $L_{hinge}$, we conduct ablation study of it with different annealing schedules, including: stop training on $L_{hinge}$ after 400 epoches (early-stop), start training on $L_{hinge}$ after 100 epoches (late-begin), and annealing $L_{hinge}$ with a constant factor 1. Fig.~\ref{fig:hingeloss_schedule} records the model performance along the training iterations. Firstly, it shows the effectiveness of the proposed $L_{hinge}$ within all the anneal schedules. Moreover, early-stop leads to a direct performance downgrade in late iterations, while late-begin performs the worst in early iterations. Annealing with a constant factor yields a similar performance as the dynamic annealing in early iterations, but falls back later when the models converge. 

\subsubsection{Language Model Performance} Apart from a strong T2I performance, TIME also yields $D$ as a well-performed stand-alone image captioning model.
\begin{table}[h]
  \begin{center}
       \resizebox{0.9\linewidth}{!}{
           \begin{tabular}{l | p{20mm} p{20mm} p{20mm}}
           \toprule
          model & Captioning \newline BLEU-4 $\uparrow$ & Image \newline Retrieval@5$\uparrow$ & Text \newline Retrieval@5$\uparrow$
            \\
            \midrule
         Bert Basic (with 24 TF) & 0.389 & 69.3 & 82.2 \\
         UNITER (with 24 TF)  & 0.395 & 76.7 & 87.0 
           \\
         OSCAR (with 24 TF)   & 0.405 & 80.8 & 91.1 \\
         TIME (with 8 TF)   & 0.381 & 75.1 & 78.2 \\
           \bottomrule
           \end{tabular}
       }
   \end{center}
   \caption{Results on downstream Vision-Language tasks from TIME on COCO, compared with SOTA models.}
   \label{table:vl_performance}
   \vspace{-2mm}
\end{table}

\begin{figure*}
    \centering
    \includegraphics[width=0.86\linewidth]{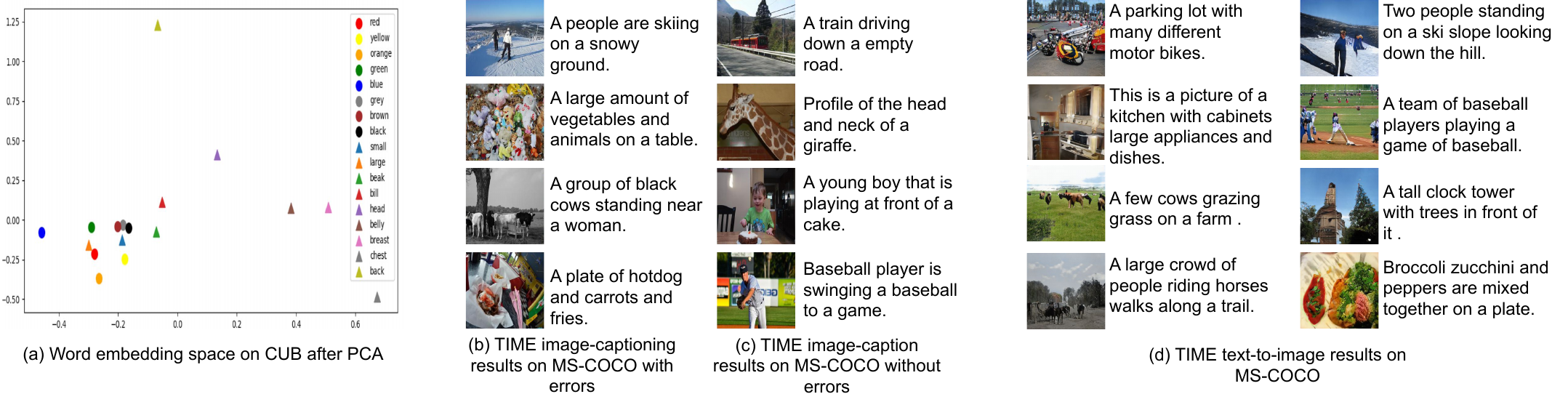}
    \caption{Learned word embeddings on CUB, and qualitative results on MS-COCO}
    \label{fig:text_embedding}
\end{figure*}

\begin{table*}[h]
  \begin{center}
    \resizebox{1.8\columnwidth}{!}{
      \begin{tabular}{l| l| c c c c c c c c c}
      \toprule
      \multicolumn{2}{c}{} & StackGAN & AttnGAN & ControlGAN & MirrorGAN  & DMGAN  & TIME & Real-Image \\
      \cmidrule{1-9}
      \multirow{2}{*}{CUB} & Inception Score $\uparrow$ & $3.82 \pm 0.06 $ & $4.36 \pm 0.03$ & $4.51 \pm 0.06$ & $4.56 \pm 0.05$ &$4.71 \pm 0.02$ &  $\textbf{4.91} \pm 0.03$ & $5.04 $        \\
      & FID $\downarrow$ & N/A & 23.98 & N/A & N/A  & 16.09 & \textbf{14.3} & 0 \\
      & R-precision $\uparrow$ & $10.37 \pm 5.88$ & $67.82 \pm 4.43$ & $69.33 \pm 3.21$ & $69.58 \pm 4.39$  & $72.31 \pm 0.91$  & $71.57 \pm 1.2 $ & N/A \\
      \cmidrule{1-9}
      \multirow{2}{*}{COCO} & Inception Score $\uparrow$ & $8.45 \pm 0.03$ & $25.89 \pm 0.47$ & $24.06 \pm 0.6$ & $26.47 \pm 0.4$ &  $30.49 \pm 0.5$ & $ \textbf{30.85} \pm 0.7$ & $ 36.5 $ \\
      & FID $\downarrow$ & N/A & 35.49 & N/A & N/A  & 32.64 &  \textbf{31.14} & 0 \\
      & R-precision $\uparrow$ & N/A & $ 83.53 \pm 0.43$ & $82.43 \pm 2.21$ & $ 84.21 \pm 0.39$  & $ 91.87 \pm 0.28 $  & $89.57 \pm 0.9 $ & N/A \\
      & SOA-C $\uparrow$ & N/A & 25.88 & 25.64 & 27.52 & \textbf{33.44} &  32.78 & 74.97 \\
      \bottomrule
      \end{tabular}
     }
 \end{center}
 \caption{Text-to-Image performance comparison between TIME and other models.}
 \label{table:compare_sota}
\end{table*}

 Table~\ref{table:vl_performance} shows the comparison between TIME and more complicated NLP models, reflects the practicality and competence of TIME on the more general Vision-Language (VL) tasks. Note that we compete with Bert \cite{dai2019transformer,Pan2020XLinearAN}, UNITER \cite{chen2019uniter} and OSCAR \cite{li2020oscar} which all are large scale models with 24 Transformer (TF) blocks, and pre-trained on multiple VL tasks for an optimal performance.

In contrast, TIME is only trained on the studied text-image mutual translation task, with a tiny model size (only 8 TF blocks) and without any pre-training. It gains close performance to the SOTA models, which reveals a promising area for future research towards mutual-translation in a single framework. Fig.~\ref{fig:text_embedding} shows the qualitative results from TIME on language tasks. In Fig.~\ref{fig:text_embedding}-(a), words with similar meanings reside close to each other. ``Large" ends up close to ``red", as the latter often applies to large birds, while ``small" is close to ``brown" and ``grey", which often apply to small birds.

\subsubsection{Comparison on T2I with State-of-the-Arts}
We next compare TIME with several SOTA text-to-image models. Qualitative results of TIME can be found in Figs.~\ref{fig:quality_1}, \ref{fig:quality_2}, and \ref{fig:text_embedding}. On CUB, TIME yields a more consistent image synthesis quality, while AttnGAN is more likely to generate failure samples. On MS-COCO, where the images are much more diverse and complex, TIME is still able to generate the essential contents that is consistent with the given text. The overall performance of TIME proves its effectiveness, given that it also provides image captioning besides T2I, and does not rely on any pre-trained modules. 

As shown in Table~\ref{table:compare_sota}, TIME demonstrates competitive performance on MS-COCO and CUB datasets with the new state-of-the-art IS and FID. Unlike the other models that require a well pre-trained language module and an Inception-v3 image encoder, TIME itself is sufficient to learn the cross-modal relationships between image and language. Regarding the image--text consistency performance, TIME is also among the top performers on both datasets. Specifically, we do not tune the model structure to get an optimal performance on MS-COCO dataset. As our text decoder in $D$ performs image captioning with image feature-map of size $8\times8$, such a size choice may not able to capture small objects in images from MS-COCO. On the other hand, $8\times8$ is the suitable size to capture features of bird parts, for images from CUB dataset. 

Importantly, TIME is considerably different from AttnGAN (no pre-training, no extra CNN/RNN modules, no stacked structure, no sentence embedding), while the other models all based on AttnGAN with orthogonal contributions to TIME. These technique contributions (e.g. DMGAN, SD-GAN, OP-GAN) could also be incorporated into TIME, with foreseeable performance boost. Due to the content limitation, we omit the integration of such ideas.

\section{Conclusion}
In this paper, we propose the Text and Image Mutual-translation adversarial nEtwork (TIME), a unified framework trained with an adversarial schema that accomplishes both the text-to-image and image-captioning tasks. Via TIME, we provide affirmative answers to the four questions we raised in Section 1. While previous works in the T2I field require pre-training several supportive modules, TIME achieves the new state-of-the-art T2I performance without pre-training. The joint process of learning both a text-to-image and an image-captioning model fully harnesses the power of GANs (since in related works, $D$ is typically abandoned after training $G$), yielding a promising Vision-Language performance using $D$. TIME bridges the gap between the visual and language domains, unveiling the immense potential of mutual translations between the two modalities within a single model.

\clearpage

{\small
\fontsize{10pt}{11pt} \selectfont
\bibliographystyle{aaai21}
\bibliography{egbib}
}
\clearpage
\appendix

{\noindent\textbf{Appendix}}

\section{Broader Impact} 
Generating realistic images from text descriptions is an important task with a wide range of real-world applications, such as reducing repetitive tasks in story-boarding, and film or video game scene editing. However, such technology may also lead to the spread of misinformation via synthesized images, and may enable infringement via massive re-production and simple manipulation of copyrighted images.

\section{Implementation Details}

The code that re-produces the results of TIME along with trained model checkpoints and pre-processed data are available at: \url{https://drive.google.com/drive/folders/1tMgEB7F7ZH0J2lfaPOsowwOJgq8ygUDT?usp=sharing}. Please check the detailed training configurations in the code. A ``readme" file is as well included to link the proposed modules to the code snippets. Our training environment consists of two NVIDIA TITAN RTX GPUs, each of which has 24 GB memory, with the CUDA-10.1 driver and PyTorch framework running on the Ubuntu-19.10 operating system. During training, not all GPU capacity is necessarily used.

\section{Vanilla Conditional Loss}
Apart from the proposed image-captioning loss, which serves as a conditioning factor to train $D$ and $G$, we also applied another simple conditional loss by combining the image feature and text embeddings (this vanilla conditional loss is reflected in Fig.~2 upper-left). Such conditional loss serves as a basic regulation term for $D$, just like the conditional loss used in StackGAN. 

In practice, we can extract $8\times8$ or $4\times4$ image features from the CNN part of $D$, and concatenate the image features with the reshaped word embeddings along the channel dimension. Specifically, we re-shape the word embeddings by fitting each of them into one pixel location of an image feature map. For example, if we have an image feature with the shape of $512\times4\times4$ ($4\times4$ spatial size with 512 channels), and a text description of 16 words, where each word is represented in $512$ dimensions, we reshape the word embeddings into $512\times4\times4$, where the top-left spatial location is the first embedding, the second top-left location is the second embedding, and the bottom-right contains the last embedding, etc. The image features and the reshaped word embeddings can then be concatenated into a image-context feature with the shape of $1024\times4\times4$. 
Such image-context feature is then processed by two convolution layers, giving rise to the final image--text matching score.

When the word sequence is longer than 16, we just select the first 16 words, and when the sequence is less than 16, we use 0-padding for the remaining space. This sort of procedure is fairly naive, and can potentially be improved. However, this is not the primary concern in this work and we find it already works sufficiently well across all of our experiments. Since using such a conditional loss is a common practice in prior works, we only conduct a simple ablation experiment on TIME and AttnGAN to confirm the effect of this loss. The results shows a consistent small performance gain for both TIME and AttnGAN, where abandoning such a conditional loss only leads to about $2\%$ IS loss on the CUB dataset. As a result, we pick this conditional loss for all models. Further details of the implementation can be found in the code.

\section{2-D Positional Encoding}
The formal equation of the 2-D positional encoding is:
\begin{small}
\begin{align}
    P_\mathrm{i\in[1:\frac{c}{4}]}(y, 2i) = \sin\left(\frac{y}{10000^{\frac{4i}{c}}}\right) ;\\
    P_\mathrm{i\in[\frac{c}{4}:\frac{c}{2}]}(x, 2i) = \cos\left(\frac{x}{10000^{\frac{2i}{c}}}\right) ;\\
    P_\mathrm{i\in[1:\frac{c}{4}]}(y, 2i-1) = \cos\left(\frac{y}{10000^{\frac{4i}{c}}}\right) ;\\
    P_\mathrm{i\in[\frac{c}{4}:\frac{c}{2}]}(x, 2i-1) = \sin\left(\frac{x}{10000^{\frac{2i}{c}}}\right),
\end{align}
\end{small}where $x$, $y$ are the coordinates of each pixel location, and $i$ is the dimension index along the channel. 

One interesting question about 2-D positional encoding is: ``if we cut off the boundary parts of the input image, which leads to major value changes of the position encoding on all pixels, how will it affect the captioning result?"

The purpose of position encoding is to re-value the feature vectors based on each spatial location, making the attention operation better at attending to the related positions. We believe the ``relative value” between each spatial location is the key and it should be invariant in cases such as the one just  described. I.e., the neighbor pixels’ \textbf{relative value difference} should not be changed even when the boundary pixels are cut off. Therefore, each embedding will still attend to its most relative embeddings that have similar embedding values, so the attention result and the resulting captions will show little change. We conducted experiments that confirmed our hypothesis, and showed that the 2D-positional-encoding based attention operation has such a ``content invariant" property.

\section{Controllability}
The reason we gain controllability with TIME is fairly intuitive. Here, we extend the discussion in the paper on "Controllable G without Sentence-Level Embedding" with more implementation details. Unlike previous T2I models, the text information is introduced to the model only on high resolution feature-maps (e.g., 32x32 and 64x64 as illustrated in Fig.~2 of the paper), thus it is unlikely to change the content shape. This intuition is validated in many GAN works, a typical example being StyleGAN, where changing the input noise at the $8\times8$ feature level leads to major changes in content, while changing the input noise at the $32\times32$ level does not change the shapes in the image at all, but instead only leads to stylistic changes such as the coloring. Our experiments also confirmed this observation: when we tested feeding text embeddings to a lower-resolution layer (e.g., $8\times8$), the model controllability indeed disappeared.

Note that no additional tweaks were used (extra loss, extra labeled data) to gain such controllability, and thus the degree of controllability is highly biased on the data. Taking the CUB dataset as an example, although the paired text description only describes how the bird looks like, there is a high correlation between the color of birds and the background, e.g., yellow birds tend to be found more in green trees and white birds more in the blue sky. Such ``background \& bird-style" correlation can easily be learned by the model without any extra guidance. As a result, the background color also changes when we change text that merely describes the bird, and hence the degree of controllability pertains more to the invariant shape of the bird but not to an invariant background.

\section{Transformer configuration exploration}
We conducted the same experiments to explore the best attention settings for the T2I task on the MS COCO dataset. Table~\ref{table:attn_coco} lists the settings we tested, where all the models are configured in the same way based on AttnGAN, except for the attention mechanisms used in $G$.
 
\begin{table}[h]
\caption{Comparison between different attention settings on MS COCO dataset}
  \begin{center}
       \resizebox{\columnwidth}{!}{
           \begin{tabular}{c c c c c c c}
           \hline
            & AttnGAN & Tf-h1-l1 & Tf-h4-l1 & Tf-h4-l2 & Tf-h4-l4 & Tf-h8-l4     
            \\
            \hline
           Inception Score $\uparrow$ & $25.89 $ & $26.58 $ & $26.42$ & $27.48 $ & $27.85 $ & $27.15 3$        
           \\
           R-precision $\uparrow$  & $83.53 $ & $86.46 $ & $88.58 $ & $89.72 $ & $89.57 $ & $88.32$ \\
           \hline 

           \end{tabular}
       }
   \end{center}
   
   \label{table:attn_coco}
\vspace{-0.3cm}
\end{table}

The results are consistent with the experiment on the CUB dataset in showing that the ``multi-head and multi-layer" (MHL) design boosts the performance. We hypothesize that the optimal number of heads and layers depends on the dataset, where the ``4-heads 2-layers" and ``4-heads-4-layers" settings are the sweet points for the CUB and COCO datasets, respectively. While the motivation and design of MHL is quite intuitive, it is still possible that the performance gain results from the increased number of parameters. This suspicion is beyond this work. On the other hand, as shown in the last two columns, making the Transformer too ``big" leads to worse performance. A possible reason is that the increase in parameters makes it harder for the model to converge, and makes it more susceptible to overfitting the training data.

\section{More on Related Work}
In this section, we want to highlight the differences between TIME and previous work.

\subsection{StackGAN as the Image Generation Backbone}
\begin{figure*}
\centering
  \includegraphics[width=0.74\linewidth]{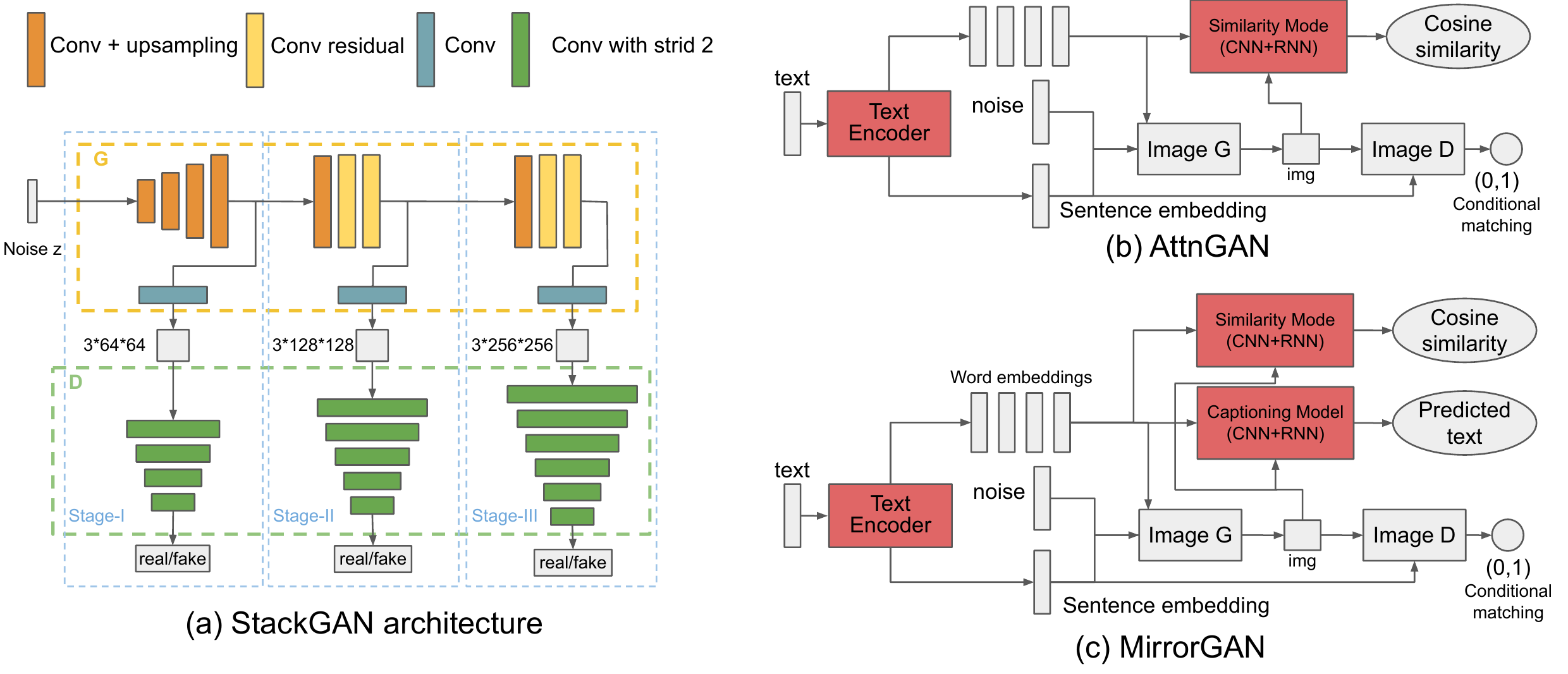}
  \caption{(a) The StackGAN structure that serves as the backbone in SOTA T2I models \cite{zhang2017stackgan,xu2018attngan,qiao2019mirrorgan,zhu2019dm,8930532,NIPS2019_8480,hinz2019semantic}. (b)\&(c) Representative models build upon StackGAN, with red parts indicating modules that require pre-training. Note that our proposed model TIME does not: 1. require pre-training the red modules in (b) and (c); 2. require multiple discriminators (the green modules) in (a). }
  \label{fig:previous_models}
\end{figure*}

StackGAN adopts a coarse-to-fine structure that has shown substantial success on the T2I task. The generator $G$ takes three steps to produce a $256\times256$ image as shown in Figure~\ref{fig:previous_models}-(a). In stage-I, a $64\times64$ image with coarse shapes is generated. In stage-II and III, the feature maps are further up-sampled to produce more detailed images with better textures. Three discriminators are required, where the lowest-resolution $D$ guides $G$ with regard to coarse shapes, while localized defects are refined by the higher-resolution $D$s.

However, there are several reasons for seeking an alternative architecture. First, the multi-$D$ design is memory-demanding and has a high computational burden during training. As the image resolution increases, the respective higher-resolution $D$s can raise the cost dramatically. Second, it is hard to balance the effects of the multiple $D$s. Since $D$s are trained on different resolutions, their learning paces diverge, and can result in conflicting signals when training $G$. In our experiments, we notice a consistently slower convergence rate of the stacked structure compared to a single-$D$ design.

\subsection{Dependence on Pre-trained modules}

While the overall framework for T2I models resembles a conditional GAN (cGAN), multiple modules have to be pre-trained in previous works. As illustrated in Figure~\ref{fig:previous_models}-(b), AttnGAN requires a DAMSM, which includes an Inception-v3 model \cite{szegedy2016rethinking} that is first pre-trained on ImageNet \cite{deng2009imagenet}, and then used to pre-train an RNN text encoder. MirrorGAN further proposes a pre-trained STREAM module as shown in Fig.~\ref{fig:previous_models}-(c).

Such pre-training has a number of drawbacks, including, first and foremost, the computational burden. Second, the additional pre-trained CNN for image feature extraction introduces a significant amount of weights, which can be avoided as we shall later show. Third, using pre-trained modules leads to extra hyper-parameters that require dataset-specific tuning. For instance, in AttnGAN, the weight for the DAMSM loss can range from 0.2 to 100 across  different datasets. While these pre-trained models boost the performance, empirical studies in MirrorGAN and StackGAN show that they do not converge if jointly trained with cGAN.

\subsection{Difference to MirrorGAN}
Our model shares a similar idea with MirrorGAN in ``matching the image captions for a better text-to-image generation consistency". However, there are substantial difference between our work and MirrorGAN. First, while MirrorGAN focused on the T2I task and uses a fixed captioning model to support the training of G, our idea can better be described as ``image--text mutual translation”. The objective is not merely to train a T2I $G$, but also to train an image captioning model at the same time. Second, the study in MirrorGAN suggests that the image-captioning module with extra CNN and RNN has trouble converging if trained together with the T2I GAN, and thus can only be pre-trained. However, in our work, we show that the image-captioning module can be jointly trained under the same framework, and the joint training leads to better performance for $G$ while does not hurt the performance on captioning for $D$. Third, we believe ``image--text mutual translation” is a topic that can attract substantial research attention and that it is particularly intriguing, as humans can easily relate visual and lingual information together.
 
TIME differs from MirrorGAN in several respects to realize the joint training. i) MirrorGAN uses LSTMs to model the language, which suffers from vanishing gradient issues, in particular that the discrete word embeddings are not very compatible with image features, which are more continuous when trained jointly. In contrast, Transformer is more robust in dealing with the combined features of the text and image modalities. ii) MirrorGAN tries to train directly on word embeddings, which is common in T2I tasks. In contrast, in our model, the word embeddings that the CNNs (G and D) take are not the initial word embeddings. We have a text encoder to endow the word sequence embeddings with contextual signals, making the resulting sentence representations much smoother, thereby facilitating training. iii) The language part of TIME is only trained with the discriminator on the image captioning task. The embeddings are not trained to help the Generator for better image quality. This focused objective frees the text encoder from potentially conflicting gradients, aiding convergence, and making mutual translation feasible with our model.

\subsection{Contribution Report}
In this paper, we explored the effectiveness of the ``Multi-Layer and Multi-Head” (MLH) attention design based on a Transformer in the T2I task. Note that its use is definitely not meant for novelty. Moreover, it is still an open problem to fully understand how and why the attention mechanism works (see e.g.\ new discussions in \cite{tay2020synthesizer}). Our aim is not to try to explore that problem, but instead we seek to know how the more comprehensive Transformer structure benefits generative image models compared to the currently used ``single-layer and single-head" attention in models such as AttnGAN and SAGAN. 

Apart from the MLH design, we would like to emphasize the ``novelty" aspects in our work as follows: we propose 1.\ a training strategy for T2I without pre-training, 2.\ a 2-D positional encoding, 3.\ annealing hinge loss for adversarial training, 4.\ a new generator structure that is more lightweight and controllable, 5.\ a ``text--image" mutual translation perspective of learning a model to bridge the image and text modalities. Importantly, the work of this paper was started in early 2019 and first submitted for peer review on March 2020 (withdrawn), one may find some of the claimed contributions not novel on the date of publication in 2021. To our best research on up-to-date related works, the integration of transformer into image models, and our aforementioned point 2. and 4. have also been studied in other works \cite{mildenhall2020nerf}, while the rest points remain unique and novel.

\section{Image Captioning Qualitative Result} 

\begin{figure}[h]
 \includegraphics[width=0.9\linewidth]{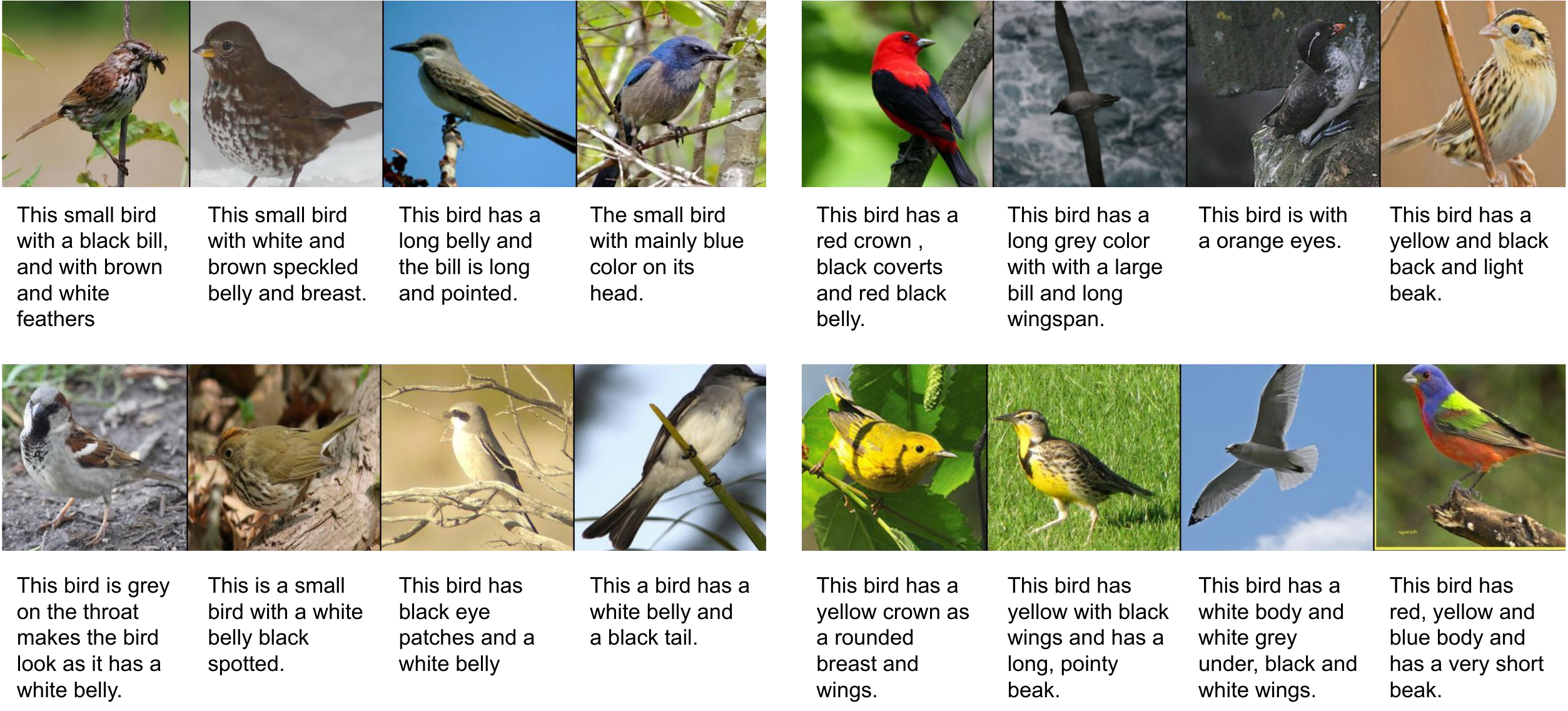}
  \caption{Image captioning results of TIME on the CUB dataset testing set.}
  \label{fig:bird_i2t}
\vspace{5mm}
  \includegraphics[width=0.9\linewidth]{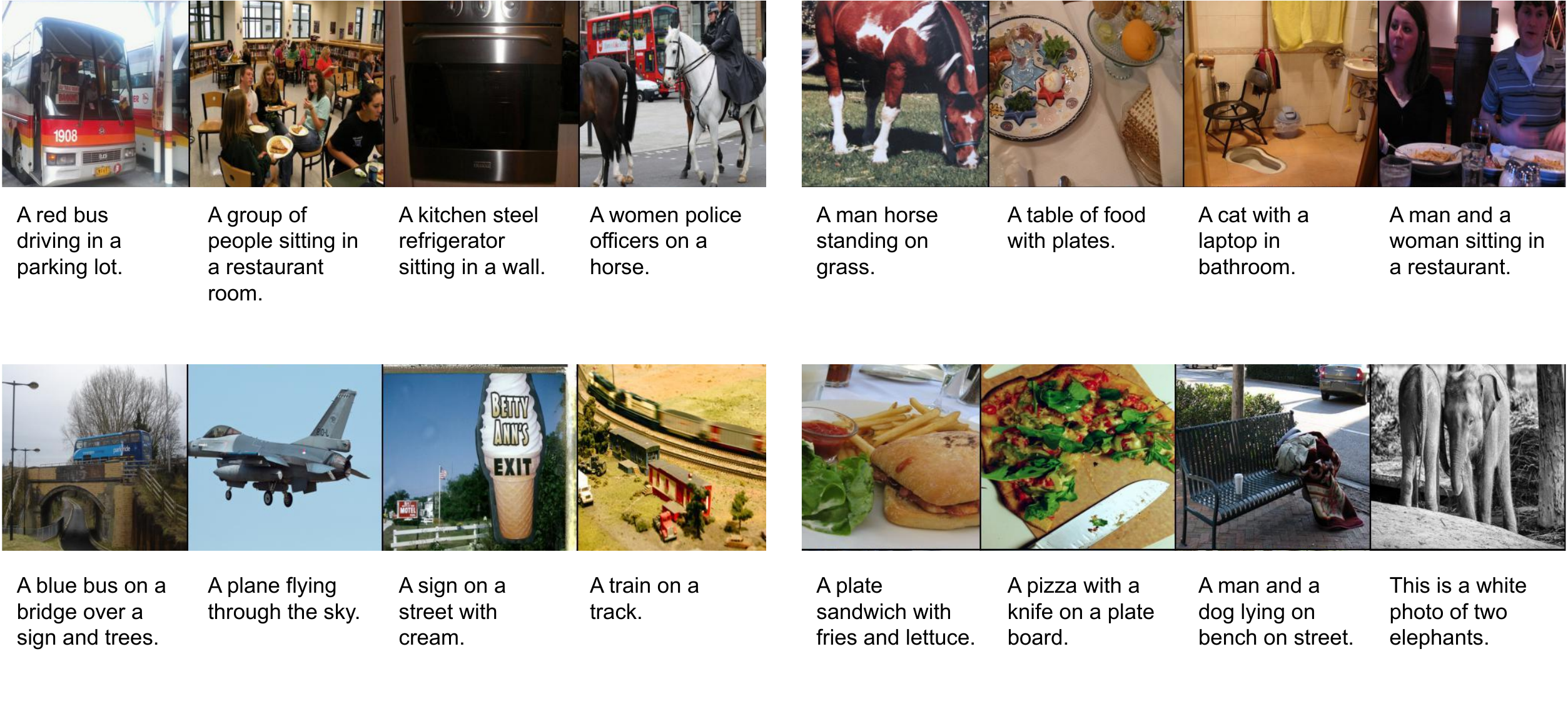}
  \caption{Image captioning results of TIME on the MS-COCO dataset testing set.}
  \label{fig:coco_i2t}
\end{figure}

\section{Text to Image Qualitative Results}

\clearpage

\begin{figure*}[!ht]
  \centering
  \includegraphics[width=0.7\linewidth]{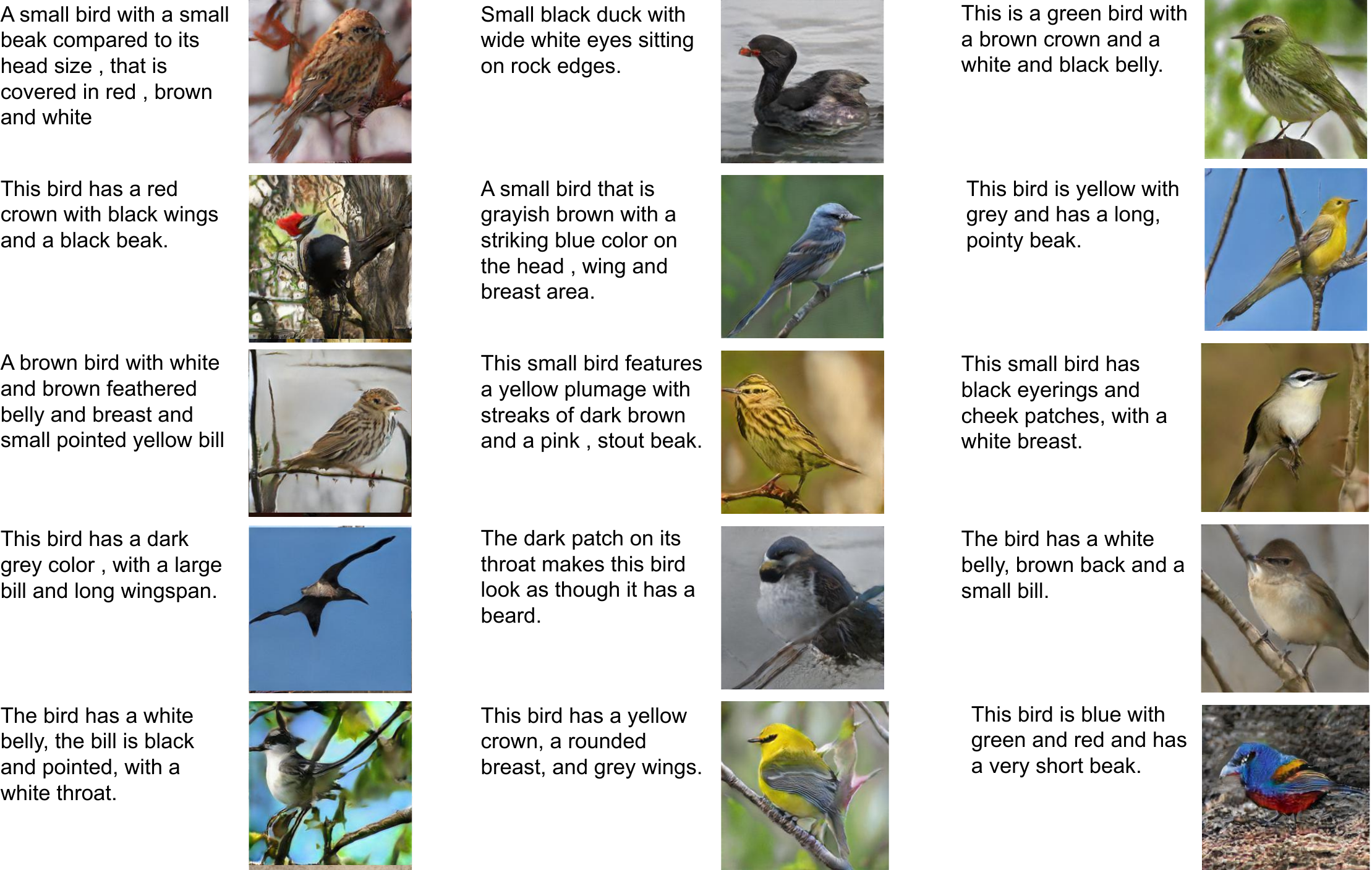}
  \captionof{figure}{Uncurated Qualitative results of TIME on the CUB dataset: The images are generated by TIME given the captions from testing dataset.}
  \label{fig:bird_i2t}
\end{figure*}

\begin{figure*}[h]
  \centering
  \includegraphics[width=0.7\linewidth]{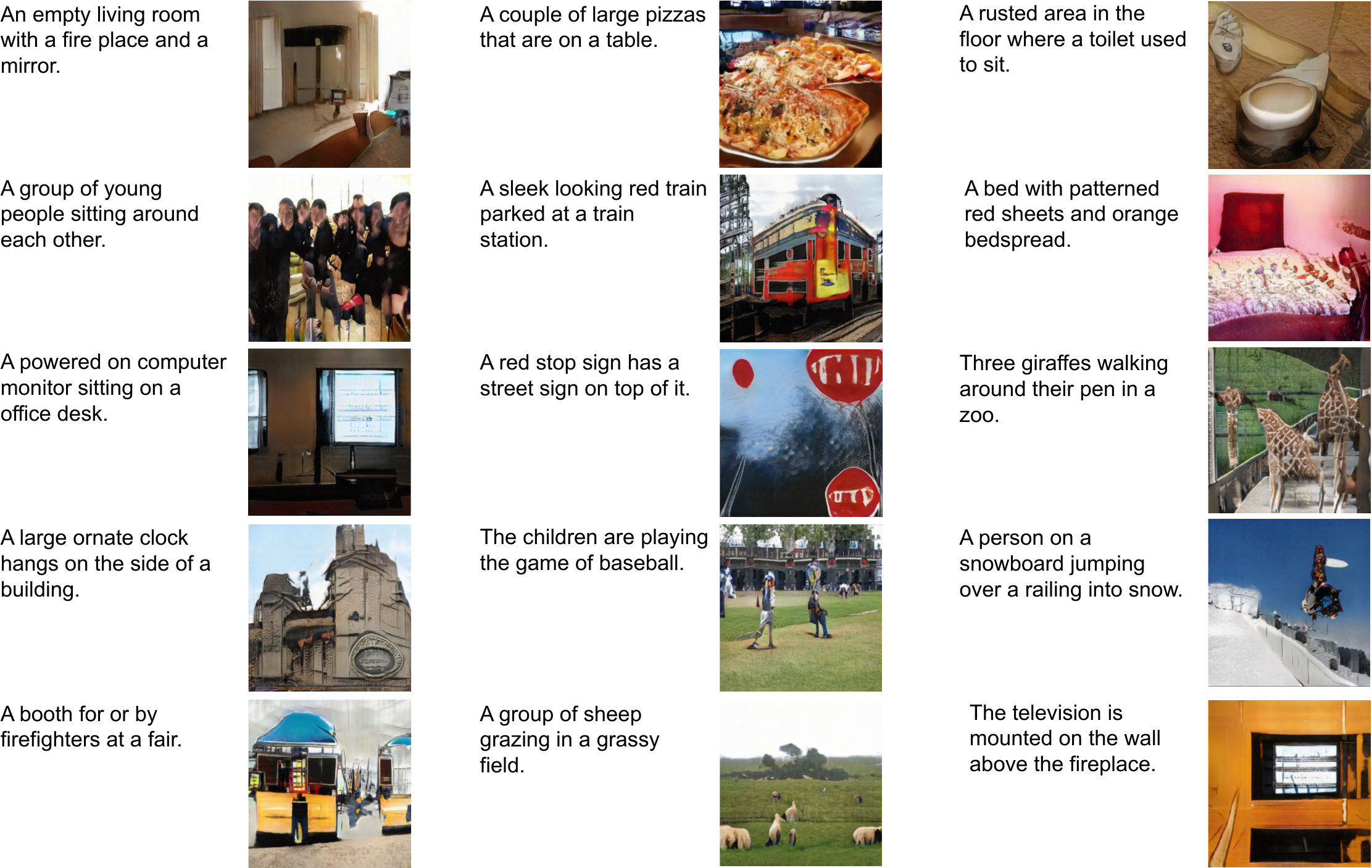}
  \captionof{figure}{Uncurated Qualitative results of TIME on the MS-COCO dataset: The images are generated by TIME given the captions from testing dataset.}
  \label{fig:coco_i2t}
\end{figure*}

\clearpage

\begin{figure*}[h]
\centering
  \includegraphics[width=0.7\linewidth]{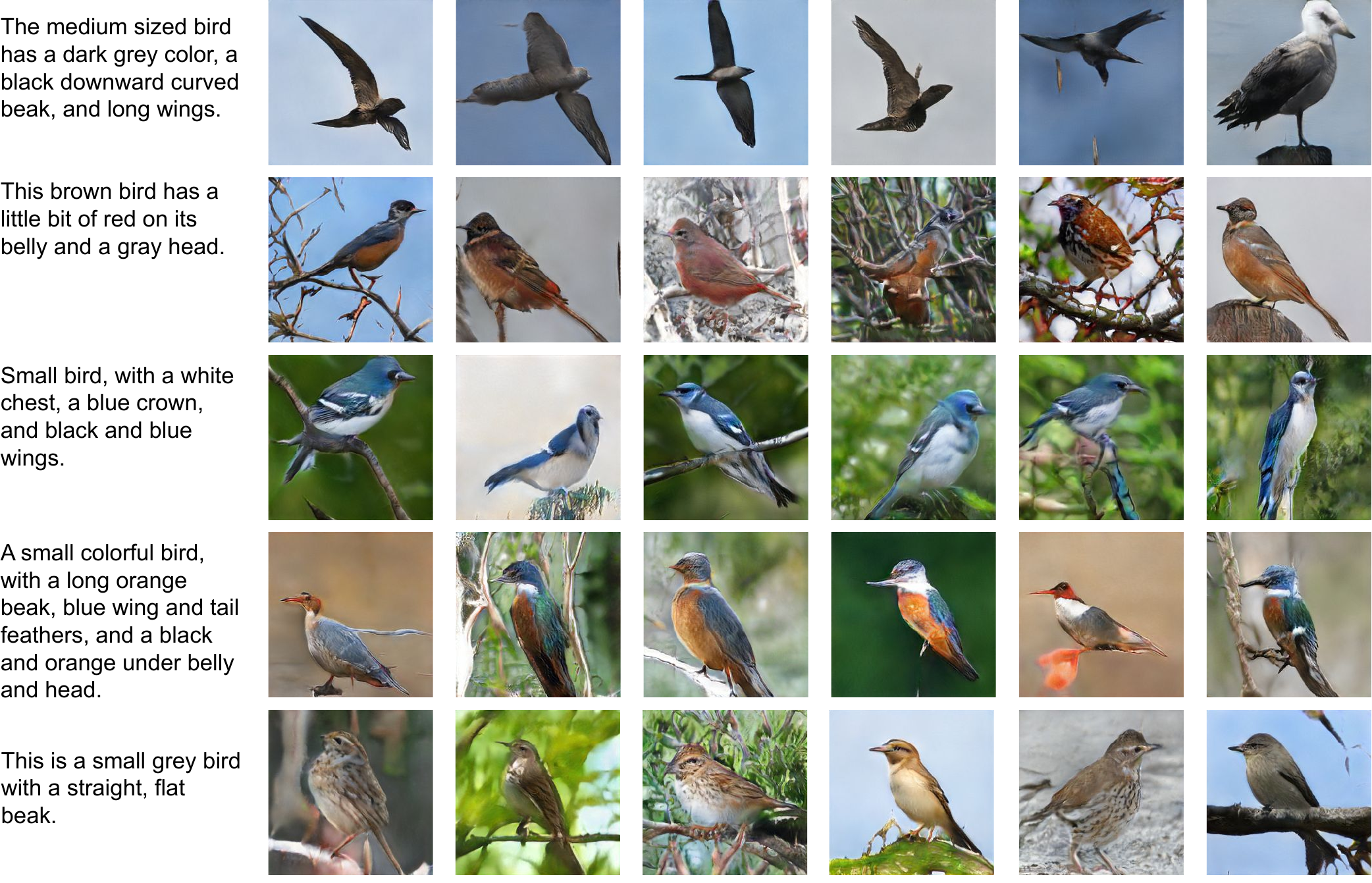}
  \caption{Uncurated Qualitative results of TIME on the CUB dataset: The images are generated by TIME given the captions from testing dataset, each row containing the generates samples from the same caption.}
  \label{fig:bird_i2t}
\vspace{5mm}
\centering
  \includegraphics[width=0.7\linewidth]{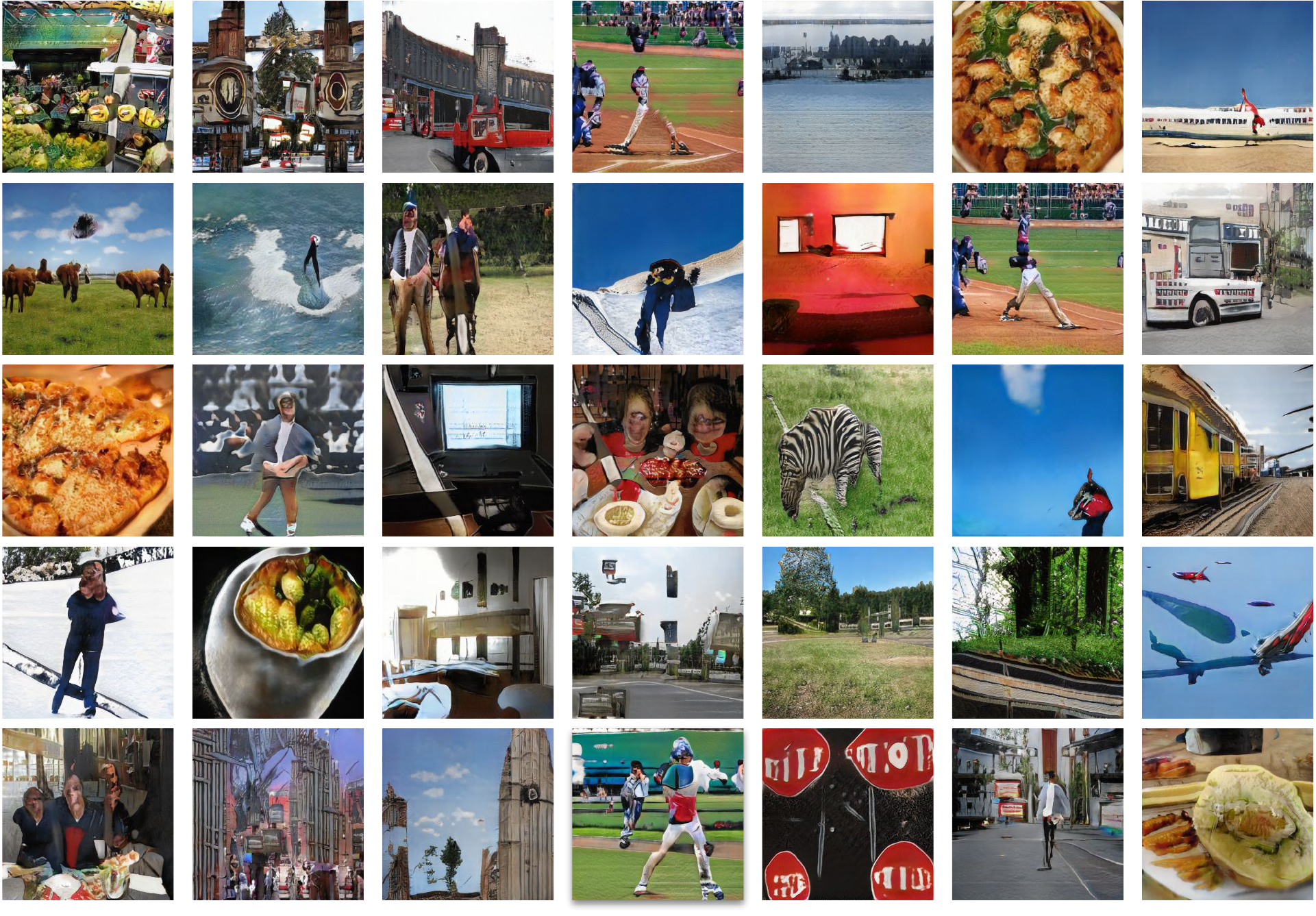}
  \caption{Uncurated Qualitative results of TIME on the MS-COCO dataset: The images are generated by TIME given the captions from testing dataset.}
  \label{fig:coco_i2t}
\end{figure*}

\end{document}